# DEPT: Deep Extreme Point Tracing for Ultrasound Image Segmentation

Lei Shi[1,2], Xi Fang[3], Naiyu Wang[1,2] and Junxing Zhang[1, *], *IEEE Member*

[1] College of Computer Science, Inner Mongolia University, Hohhot, China.
[2] Baotou Medical College, Baotou, China.
[3] Independent researcher
Email: shilei@mail.imu.edu.cn; junxing@imu.edu.cn

*Abstract*—Automatic medical image segmentation plays a crucial role in computer aided diagnosis. However, fully supervised learning approaches often require extensive and labor-intensive annotation efforts. To address this challenge, weakly supervised learning methods, particularly those using extreme points as supervisory signals, have the potential to offer an effective solution. In this paper, we introduce Deep Extreme Point Tracing (DEPT) integrated with Feature-Guided Extreme Point Masking (FGEPM) algorithm for ultrasound image segmentation. Notably, our method generates pseudo labels by identifying the lowest-cost path that connects all extreme points on the feature map-based cost matrix. Additionally, an iterative training strategy is proposed to refine pseudo labels progressively, enabling continuous network improvement. Experimental results on two public datasets demonstrate the effectiveness of our proposed method. The performance of our method approaches that of the fully supervised method and outperforms several existing weakly supervised methods.

*Keywords—medical image segmentation, weakly supervised learning, Deep Extreme Point Tracing, Feature-Guided Extreme Point Masking*

## I. Introduction

Automatic medical image segmentation, as a crucial component in medical image analysis, which plays an important role in computer aided diagnosis and treatment planning [1,2]. For example, accurately determining the size and boundary of the lesion or tumor in ultrasound images assists surgeons in developing the treatment plans and monitoring the progression of the disease over time, such as conditions like breast tumors and cancer [3,4].

Deep learning has dominated medical image segmentation for recent years and achieved high accuracy on different tasks [5, 6, 7]. However, despite the success of deep learning, it typically relies on a large amount of finely annotated data for training. However, pixel-wise annotation, such as delineating lesion regions, are both time consuming and labor intensive and often requires the expertise of experienced physicians. To alleviate the burden of extensive manual annotation, various types of geometrical weak annotations have been utilized to reduce the manual labor in data annotation. For example, scribbles, bounding boxes and extreme points have been used to generate pseudo labels to supervise the training of deep neural networks.

Extreme points, i.e., topmost, leftmost, bottommost, and rightmost points, can provide additional geometrical information and can be easily obtained during the standard process of mask and box annotation. The existing extreme points-based segmentation can be categorized into two categories. In the first category, extreme points are used as additional information to guide the network through interactive segmentation [8,9]. In the second category, points are used to supervise the network.

Several methods using extreme points for weakly medical image segmentation are proposed, such as random walker [10] and CRF-based method for creating pseudo masks [11]. However, they are either inaccurate or time-consuming.

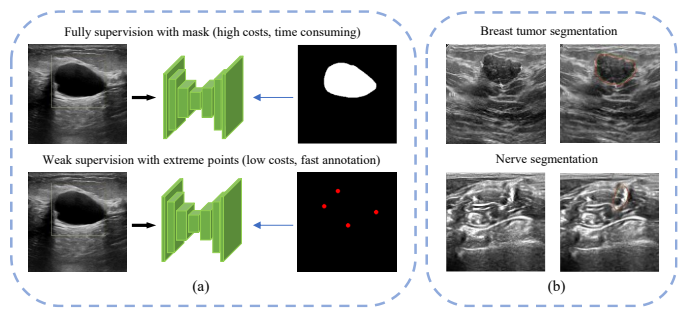

Fig. 1. (a) mask supervised method versus extreme points supervised method; (b) two visualization samples with our DEPT. left: original images, right: segmentation results. The ground truths are in green and ours are in red. Best zoomed in and viewed in color.

In this paper, we propose a novel method, namely Deep Extreme Point Tracing (DEPT), to connect four extreme points to form the initial pseudo label and an iterative training strategy to train the neural network. More specifically, we use the Sobel operator to obtain an enhanced gradient map on the input image, and use the Dijkstra's algorithm to find the shortest paths between the extreme points on the gradient map to generate initial pseudo labels that can guide the network training. With this approach, a coarse contour of the lesion region can be obtained using only four extreme points without any segmentation mask information. Meanwhile, we use iterative training for network to achieve self-strengthening of the model. During the training process, we correct the output feature map of the network using Feature-Guided Extreme Point Masking (FGEPM) for every fixed epochs to update the pseudo label of the previous stage. Through the iterative training approach, the network can eventually produces more accurate predictions than supervised only by unchanged pseudo labels. Fig.1 shows the motivation and segmentation results of our method.

The main contributions of our work are as follows:

(1) To the best of our knowledge, we propose DEPT, which is the first study that automatically create a pseudo mask from extreme points by connecting the points and filling the mask for ultrasound image segmentation.

(2) We propose a Feature-Guided Extreme Point Masking (FGEPM) algorithm, a method that dynamically refines pseudo labels generated from extreme points, thereby progressively enhancing the network's performance in return.

(3) With only four extreme points of annotation for each ultrasound image, the segmentation performance of our method on two publicly available datasets exceeds that of several state-of-the-art weakly supervised methods. Surprisingly, the performance of our method is also very close to fully supervised manner for breast tumor segmentation, further validating the effectiveness of our proposed method.

## II. RELATED WORKS

### A. Weakly Geometry-Supervised Image Segmentation

Weakly supervised image segmentation aims to segment images with limit annotations, such as bounding boxes, extreme points and scribbles. Among them, methods using the bounding boxes account for a large proportion. W. Li et al. [12] use level set evolution for natural image instance segmentation. J. Wang and B. Xia [13] integrate a multiple instance learning strategy based on polar transformation to assist image segmentation. For medical image segmentation, J. Wei et al. [14] propose the mask-to-box (M2B) transformation and a scale consistency (SC) loss to improve polyp segmentation performance. Several methods using scribbles are also proposed. For example, Q. Chen and Y. Hong [15] propose a scribble-based volumetric image segmentation architecture with a label propagation module. However, bounding boxes enclose entire rectangular areas, which will inevitably introduce background noise to labels. Scribbles are hand-drawn rough lines that lack precise boundary information. Meanwhile, both of them require longer annotation time and larger cost than extreme points. Therefore, we choose extreme points as the annotation for weakly supervised learning in this paper, especially for target with convex or near-convex shapes.

### B. Extreme points for image segmentation

Extreme points for segmentation can generally be divided into two categories. In the first category, extreme points are adopted as additional information to guide the segmentation. K.K. Maninis et al. [8] use extreme points as an additional channel input to the CNN, and annotate an extra point in the erroneous area for inaccurate case to recover performance. Z. Wang et al. [9] make interactive annotation by incorporating user clicks on the extreme boundary points. However, the above methods have high model complexity and it is challenging to transfer these methods to medical tasks.

In the second category, extreme points are typically used to supervise the network for training. H. Lee et al. [16] propose a point retrieval algorithm for natural image instance segmentation using extreme points. The method in [10] uses pseudo-mask generation with random walker algorithm and minimal user interaction for medical image segmentation. R. Dorent et al. [11] use deep geodesics to connect extreme points and employ a CRF regularized loss for 3D Vestibular Schwannoma segmentation.

## III. PROPOSED METHOD

Different from the above methods, our method has both advantages for interactive segmentation and weak supervision. In this section, we present Deep Extreme Point Tracing (DEPT), which progressively update the pseudo labels by connecting the extreme points for supervising the segmentation network. Fig.2 shows the framework of the proposed DEPT method, which consists of three major modules: 1) feature extraction for cost matrix construction, and 2) pseudo-label generation through Feature-Guided Extreme Point Masking (FGEPM), and 3) iteratively mask generation for progressive training. The details of the three components are outlined as follows:

### A. Feature extraction for cost matrix establishment

The DEPT method begins by utilizing the UNeXt network [3], a lightweight and efficient convolutional architecture, as the feature extractor to extract segmentation-wise features from the input data, namely features before the sigmoid layer. Based on the extracted features, a cost matrix is established to model the spatial relationships between each pixel in the image.

To establish the cost matrix, the image is represented as a graph $G = (V, E)$, where each pixel is transformed into a vertex $v \in V$, and the edges $e \in E$ connect each vertex to its 8 adjacent neighbors. The gradient at each pixel is computed using the Sobel operator, capturing edge information and high-level feature transitions. To construct the cost matrix $C$, the reciprocal of the gradient magnitude is used to define the edge weights, capturing the relationship between vertices. Then the reciprocal of the gradients are used as the values in the cost matrix. The cost for moving between adjacent vertices $v_i$ and $v_j$ is given by:

$$C(v_i, v_j) = \frac{1}{\sqrt{G_x(i,j)^2 + G_y(i,j)^2 + \epsilon}} \quad (1)$$

These values are then used to define the edge weights of the graph, effectively encoding the connectivity and boundary information. This matrix is used to identify and connect the four extreme points of the object or region of interest.

### B. FGEPM for pseudo label generation

To delineate the region of interest, four extreme points—topmost $P_{top}$, bottommost $P_{bot}$, leftmost $P_{left}$, and rightmost $P_{right}$—are identified as anchors. Since the original dataset generally does not provide extreme points annotation, we need to extract four extreme points consistent with the annotator from the ground truth. It is worth noting that the ground truth will be no longer used in all the next steps to strictly follow the weakly supervised learning form.

In order to effectively connect extreme points to generate pseudo label, we propose a Feature-Guided Extreme Point Masking (FGEPM) algorithm. More specifically, with extreme points, the optimal paths between consecutive pairs are computed by applying Dijkstra's algorithm on the gradient map.

For any two adjacent extreme points $P_a$ and $P_b$, the shortest path is computed as :

$$S(P_a, P_b) = \arg\min_{\mathcal{P}} \sum_{(v_i, v_j) \in \mathcal{P}} C(v_i, v_j) \qquad (2)$$

where $\mathcal{P}$ represents all possible paths between $P_a$ and $P_b$. The four shortest paths $S(P_{top}, P_{left})$, $S(P_{left}, P_{bot})$, $S(P_{bot}, P_{right})$, $S(P_{right}, P_{top})$ are connected sequentially to form a closed contour, where the interior is assigned a value of 1 (foreground) and the exterior is assigned 0 (background). This algorithm ensures that the pseudo labels align with the inherent structure of the feature map. To show the main flow of the FGEPM algorithm more clearly, we list the key steps of the algorithm in Table 1.

To improve efficiency and minimize memory usage, the input image and gradient map are downsampled (e.g., using a scale factor of 0.5). After generating the lesion contour on the downsampled gradient map, bilinear interpolation is applied to upsample the result to its original resolution.

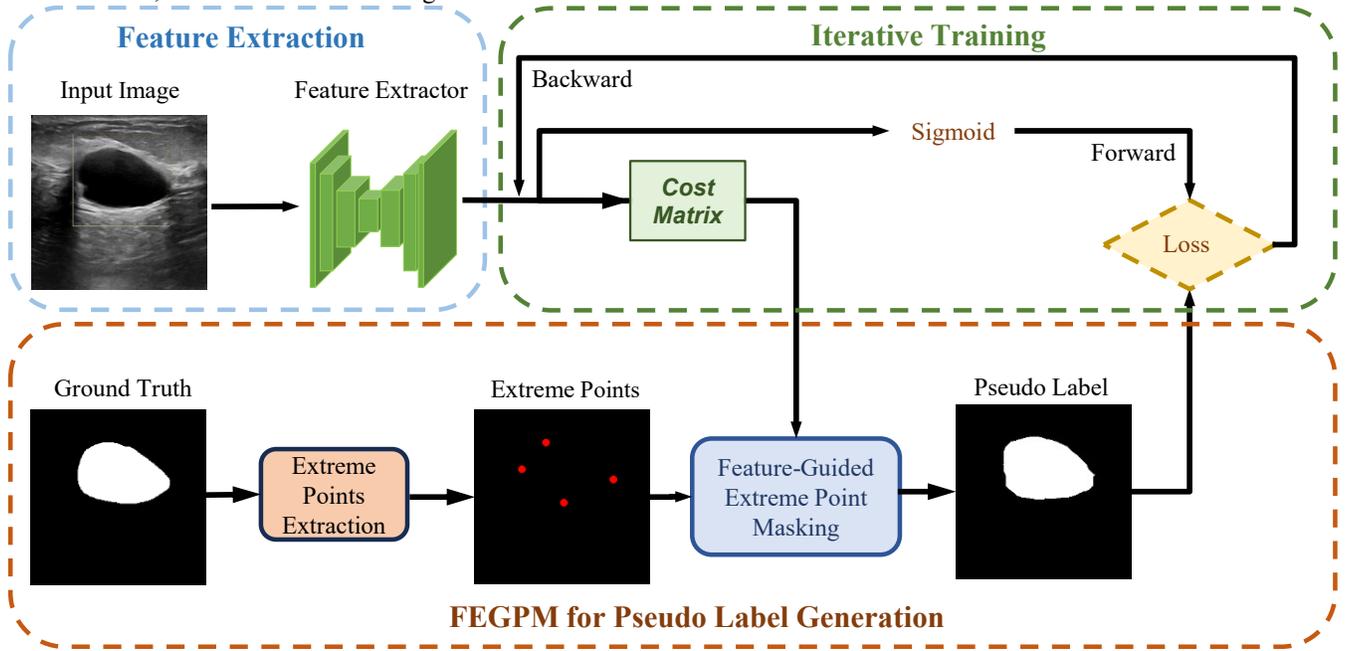

Fig. 2. The framework of our DEPT method.

TABLE I. FEATURE-GUIDED EXTREME POINT MASKING ALGORITHM

| Algorithm 1: Feature-Guided Extreme Point Masking |
| --- |
| **input:** feature map |
| **output:** pseudo label |
| 1 gradient_map = Sobel(featuremap) |
| 2 cost_matrix = (1/(gradient_map + 1e-5)) |
| 3 points = [extreme_points(topmost, bottommost, leftmost, rightmost)] |
| 4 **for** i =1,2,3,4 **do** |
| 5     route, _ = route_through_array(cost_matrix, points[i], points[i + 1], fully_connected=True) |
| 6     routes.append(route) |
| 7 **end** |
| 8 connect all routes as lesion contour |
| 9 fill interior area of the contour to generate pseudo label |

### C. Iteratively mask generation for progressive training

At the start of training, since the network is uninitialized and no feature map is available, the gradient map is directly computed from the input image to construct the cost matrix and generate the initial pseudo label. To enhance the clarity and contrast of the gradient map, Contrast Limited Adaptive Histogram Equalization (CLAHE) is applied. This step makes lesion contours more distinct, enabling more accurate connections between extreme points.

However, these initial pseudo labels often differ significantly from the ground truth, which can hinder segmentation performance if used throughout the training process. To address this, we propose an iterative training strategy that leverages the FGEPM algorithm to refine pseudo labels as the network trains. Specifically, during the forward pass, the original medical images are input into a U-shaped network to generate segmentation-wise features. From these features, two outputs are produced: (1) prediction probability maps after the sigmoid layer and (2) pseudo labels derived from the segmentation-wise feature map and extreme points using the FGEPM algorithm. A loss function is then computed based on the difference between the prediction probability maps and the pseudo labels. During the backward pass, the network minimizes this loss function, progressively improving the prediction maps and pseudo labels.

To balance efficiency and accuracy, the pseudo labels are not updated after every epoch. Instead, they are periodically updated every $n$ epochs, where $n$ is a predefined interval. This periodic update ensures that the network is supervised with increasingly accurate pseudo labels, improving segmentation performance without introducing excessive computational overhead. We totally train the neural network for $N$ epochs. At the end of the iterative training process, the optimized network weights are saved and evaluated on the test sets of various datasets, demonstrating the effectiveness of the proposed method.

## IV. EXPERIMENTS AND RESULTS

### A. Datasets

Two datasets are used to evaluate the effectiveness of our method, including Breast UltraSound Images (BUSI) [17] dataset and Ultrasound Nerve Segmentation (UNS) [18]. The original BUSI dataset contains normal, benign, and malignant images with corresponding segmentation masks. Following [3],we choose only benign and malignant image pairs. Nerve structures need to be identified in the UNS dataset to insert a patient's pain management catheter. Consistent with [19],we use 2,323 ultrasound image and mask pairs for binary segmentation tasks.

All the images and masks are resized to the resolution of 256*256, and we set a 8:2 partition for training and testing. The original images in both ultrasound dataset are grayscale, and the target areas in some images exhibit intensity distributions that are highly similar to those of the background, making it difficult to distinguish boundaries accurately. Especially for the UNS dataset, it's difficult even for physicians to find the exact location of the nerves. The above data properties pose significant challenges for designing robust segmentation methods, especially weakly supervised ones.

### B. Implementation Details

In our experiments, we set $N$ to be 400 and $n$ to be 50.Namely, we train the network for totally 400 epochs and every 50 epochs we update the pseudo labels. Note that during training, our method relies solely on extreme point annotations. During testing, the segmentation masks (ground truths) given by original dataset are then used to evaluate the segmentation performance on the test set. A single Nvidia A100 GPU is used for our experiments. We basically follow UNeXt [3] for other settings not mentioned here.

### C. Evaluation Metrics

To assess the segmentation performance of different methods, we use Intersection over Union (IoU) and Dice coefficient (also known as *F1* score) to measure the similarity between the predicted masks and the ground truth. We calculate the IoU and Dice for each image, and then average these values over the entire test set to report the mean IoU (mIoU) and mean Dice (mDice). The IoU and Dice are calculated as follows, respectively:

$$IoU = \frac{TP}{TP + FN + FP} \quad (3)$$

$$Dice = \frac{2TP}{2TP + FN + FP} \quad (4)$$

### D. Comparison with other methods

Table 2 shows the quantitative comparison of our proposed method with some existing methods. Bounding boxes and fully supervised refers to utilize bounding box labels and ground truth of trainset to supervise the training , respectively. Undoubtedly, the fully supervised approach has the best segmentation performance, which can be regarded as the upper bound of different methods. Bounding boxes approach has lower segmentation performance because it introduce background noise into the boxes. Meanwhile, we select two existing weak supervision methods for comparison: BoxInst [20] based on bounding boxes and Deep geodesics [11] based on extreme points. Note that the original baselines were either designed for the instance segmentation task or applicable to 3D medical image segmentation. Therefore, they have been adjusted accordingly. BoxInst supervises the horizontal and vertical projections of the predicted map with the bounding box annotation to minimize the gap between the tightest box of the predicted map and the ground truth. Deep geodesics are generated to connect six extreme points of a 3D target. For fair comparison, we modify the implementation of deep geodesics to fit the task of 2D medical image segmentation.

As can be seen from Table 2, the segmentation performance of our method on both datasets is closest to the fully supervised performance. At the same time, the performance of our method also surpasses other bounding box and extreme point based methods, further verifying the effectiveness of our method. This is because bounding box-based methods will inevitably guide the shape of the prediction map to a square-like shape. Although other extreme point-based methods can connect extreme points to generate initial pseudo labels, they fail to update the pseudo labels in time to provide more precise supervision for the training process.

TABLE II.   QUANTITATIVE RESULTS EVALUATED ON DIFFERENT DATASETS.

| Dataset / Method | BUSI | | UNS | |
|---|---|---|---|---|
| | *IoU* | *Dice* | *IoU* | *Dice* |
| **Bounding boxes** | 0.6052 | 0.7204 | 0.5747 | 0.7155 |
| **BoxInst [20]** | 0.6319 | 0.7336 | 0.6271 | 0.7545 |
| **Deep geodesics [11]** | 0.4678 | 0.5904 | 0.4818 | 0.6359 |
| **Fully supervised** | 0.6928 | 0.7834 | 0.6703 | 0.7870 |
| **Ours** | 0.6891 | 0.7830 | 0.6455 | 0.7698 |

## E. Ablation study

To determine the effectiveness of each component of our proposed method, we perform ablation study as well. The results are shown in Fig.3. Since the four extreme points cannot be directly used as labels to guide the training of the network, we use a simple connection of the four extreme points (e.g., two by two straight lines in counterclockwise direction) as the basic approach. Seen from Fig.3, if we connect extreme points based on gradient map to generate pseudo labels, IoU boosts by 5.9 pp (percentage point) for BUSI. If we continue to update pseudo labels in an iterative training way, IoU further increases by 5.9 pp. The same trend is shown for the UNS dataset. The above results show the positive effect of each component, as well as the overall effectiveness of our method.

## F. Visualization of results

Fig.4 shows the qualitative comparison results between prediction maps with different methods on several images from both test sets. As illustrated in Fig.4, the prediction maps produced by our method are closest to the ground truths, even better than the fully supervised method. This is probably because our method regularly updates pseudo labels based on four extreme points, while the fully supervised approach does not use iterative training, which may cause the outline of the predicted results deviating from the four extreme points, leading to inaccurate prediction results. Meanwhile, it can also be observed that other methods tend to over-segment and produce more false positives. In contrast, our method can produce prediction maps that are closer to the ground truths by generating relatively accurate initial pseudo labels and regularly updating the pseudo labels with extreme points during training.

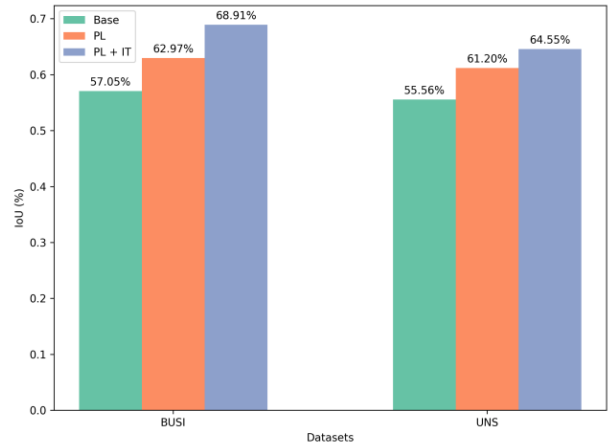

Fig. 3. IoU Comparison for BUSI and UNS datasets. Base denotes the basic approach, PL denotes the full supervised training with pseudo labels. IT denotes the iterative training approach.

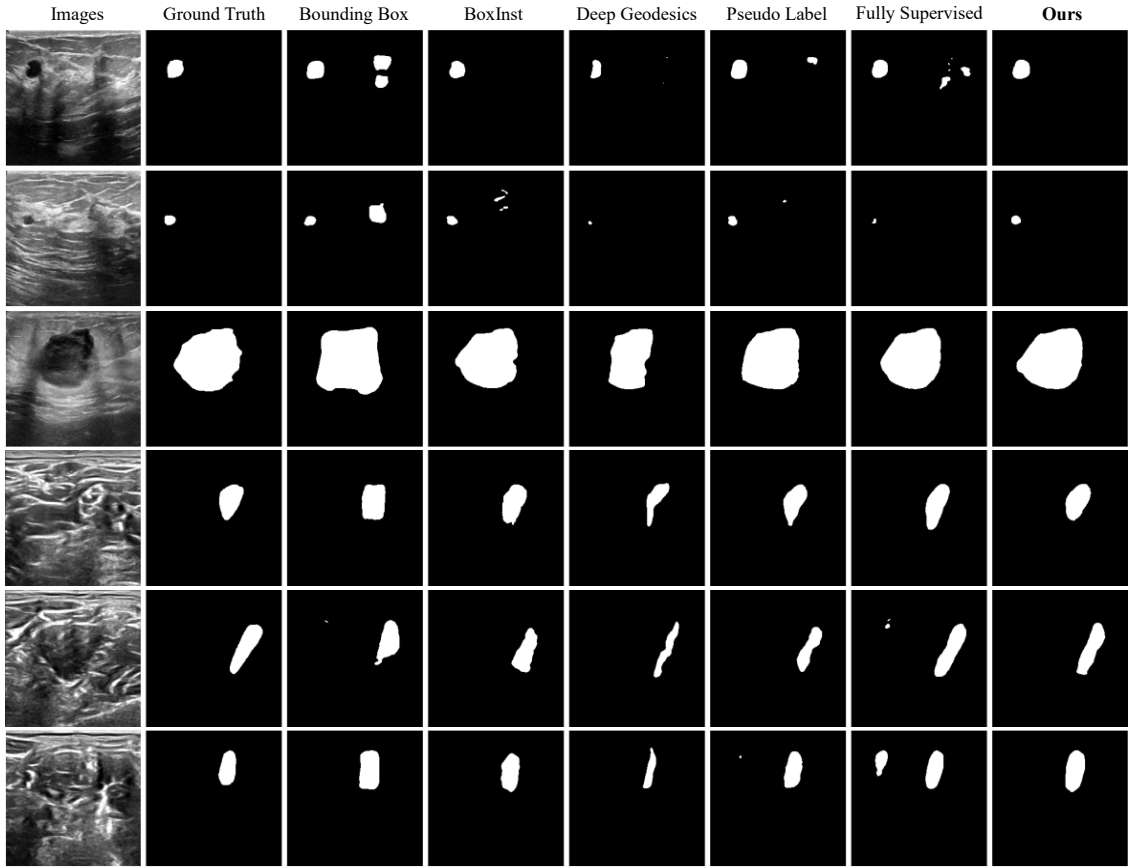

Fig. 4. Visualization comparison between prediction maps with different methods.

## V. DISCUSSION

Although our method achieves good segmentation performance on most of the images in both datasets, there are some bad examples, as shown in Fig.5. The first row of Fig. 5 is taken from BUSI, which only segments part of the tumor region. In contrast, the fully-supervised method fails to detect any foreground pixels. The second row is taken from UNS, and our method tends to detect some false positives. It is worth noting that we invite physicians to analyze the original images of the given examples, and they gave similarly large discrepancies between the segmentation results and the ground truths. The above analysis highlights the importance of developing more robust segmentation algorithms for ultrasound images.

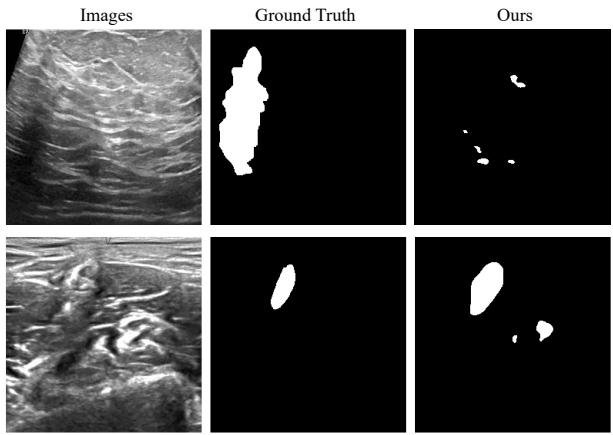

Fig. 5. Two bad examples.

Undoubtably, the quality of initial pseudo labels has a direct impact on the segmentation accuracy. In future, we will explore more algorithms for connecting extreme points to improve the quality of initial pseudo labels. We will also investigate our method on more medical datasets, especially on 3D ones. Furthermore, The loss function can also be redesigned to further enhance the weakly supervised segmentation performance.

## VI. CONCLUSION

In this paper, we propose a novel weakly supervised medical image segmentation method with only four extreme points annotation of the target. We connect the four extreme points based on gradient map from original image using Dijkstra's algorithm, in order to generate initial pseudo labels. Then, we train a U-shaped network supervised by initial pseudo labels to get prediction maps. We regularly refine and update pseudo labels in an iterative training manner, based on extreme points and the prediction maps output by the network. Both quantitative and qualitative experimental results demonstrate the effectiveness of our proposed method. Our approach eliminates the need for full pixel-level annotations while maintaining strong segmentation performance. Our approach introduces new perspectives for low-cost and high-quality weakly supervised medical image segmentation.